\documentclass[11pt,letterpaper]{article}
\usepackage[letterpaper]{geometry}
\usepackage{acl2012}
\usepackage{enumitem}
\usepackage{array}
\usepackage{hhline}
\usepackage{tabulary}
\usepackage{times}
\usepackage{xtab,afterpage}
\usepackage{latexsym}
\usepackage{amsmath}
\usepackage{eurosym}
\usepackage{mathtools}
\usepackage{amssymb}
\usepackage[usenames, dvipsnames]{color}
\usepackage{multirow}
\usepackage{url}
\usepackage[utf8]{inputenc}
\usepackage[german,russian,finnish,english]{babel}
\usepackage{cleveref}

\newcolumntype{K}[1]{>{\centering\arraybackslash}p{#1}}

\newcommand{\specialcell}[2][c]{%
  \begin{tabular}[#1]{@{}c@{}}#2\end{tabular}}

\crefname{section}{§}{§§}
\Crefname{section}{§}{§§}

\makeatletter
\newcommand{\@BIBLABEL}{\@emptybiblabel}
\newcommand{\@emptybiblabel}[1]{}
\makeatother
\usepackage[hidelinks]{hyperref}

\long\def\/*#1*/{}

\newcommand{\mcn}{\multicolumn}
\newcommand{\ti}{\textit}
\newcommand{\tb}{\textbf}
\newcommand{\mb}{\mathbf}

\newcommand{\nrr}{\color{black}}

\newcommand{\rv}{\overrightarrow}
\newcommand{\lv}{\overleftarrow}
\newcommand{\ra}{\rightarrow}

\newcommand{\clr}{\color{Red}}
\newcommand{\clg}{\color{ForestGreen}}
\newcommand{\cly}{\color{YellowOrange}}
\newcommand{\clb}{\color{Cerulean}}

\newcommand{\fenc}{\rv{f_\text{enc}}}
\newcommand{\benc}{\lv{f_\text{enc}}}
\newcommand{\dec}{f_\text{dec}}

\newcommand{\fscore}{\text{score}}
\newcommand{\fexp}{\text{exp}}

\newcommand{\foutk}{\text{out}_k}
\newcommand{\foutkp}{\text{out}_k()}
\newcommand{\foutj}{\text{out}_j}

\newcommand{\srcemb}{E_x}
\newcommand{\trgemb}{E_y}

\title{Fully Character-Level Neural Machine Translation \\ without Explicit Segmentation}

\author{Jason Lee\thanks{\hspace{1.5mm}The majority of this work was completed while the author was visiting New York University.}\\
  ETH Z\"urich \\
  {\tt\small\href{jasonlee@inf.ethz.ch}{jasonlee@inf.ethz.ch}} \\ \And
  Kyunghyun Cho \\
  New York University \\
  {\tt\small\href{kyunghyun.cho@nyu.edu}{kyunghyun.cho@nyu.edu}} \\ \And
  Thomas Hofmann \\
  ETH Z\"urich \\
  {\tt\small\href{thomas.hofmann@inf.ethz.ch}{thomas.hofmann@inf.ethz.ch}}
  }

\date{}

\begin{document}
\maketitle
\begin{abstract}
    Most existing machine translation systems operate at the level of words, relying on explicit segmentation to extract tokens. We introduce a neural machine translation (NMT) model that maps a source character sequence to a target character sequence without any segmentation. We employ a character-level convolutional network with max-pooling at the encoder to reduce the length of source representation, allowing the model to be trained at a speed comparable to subword-level models while capturing local regularities. Our character-to-character model outperforms a recently proposed baseline with a subword-level encoder on WMT'15 DE-EN and CS-EN, and gives comparable performance on FI-EN and RU-EN. We then demonstrate that it is possible to share a single character-level encoder across multiple languages by training a model on a many-to-one translation task. In this multilingual setting, the character-level encoder significantly outperforms the subword-level encoder on all the language pairs. We observe that on CS-EN, FI-EN and RU-EN, the quality of the multilingual character-level translation even surpasses the models specifically trained on that language pair alone, both in terms of BLEU score and human judgment.
\end{abstract}


\section{Introduction}\label{sec:intro}
Nearly all previous work in machine translation has been at the level of words. Aside from our intuitive understanding of word as a basic unit of meaning~\cite{Jackendoff:90}, one reason behind this is that sequences are significantly longer when represented in characters, compounding the problem of data sparsity and modeling long-range dependencies. This has driven NMT research to be almost exclusively word-level~\cite{Bahdanau:15,Sutskever:14}.

Despite their remarkable success, word-level NMT models suffer from several major weaknesses. For one, they are unable to model rare, out-of-vocabulary words, making them limited in translating languages with rich morphology such as Czech, Finnish and Turkish. If one uses a large vocabulary to combat this~\cite{Jean:15}, the complexity of training and decoding grows linearly with respect to the target vocabulary size, leading to a vicious cycle.

To address this, we present a fully character-level NMT model that maps a character sequence in a source language to a character sequence in a target language. We show that our model outperforms a baseline with a subword-level encoder on DE-EN and CS-EN, and achieves a comparable result on FI-EN and RU-EN. {\nrr A purely character-level NMT model with a basic encoder was proposed as a baseline by \newcite{Luong:16}, but training it was prohibitively slow.} We were able to train our model at a reasonable speed by drastically reducing the length of source sentence representation using a stack of convolutional, pooling and highway layers.

One advantage of character-level models is that they are better suited for multilingual translation than their word-level counterparts which require a separate word vocabulary for each language. We verify this by training a single model to translate four languages (German, Czech, Finnish and Russian) to English. Our multilingual character-level model outperforms the subword-level baseline by a considerable margin in all four language pairs, strongly indicating that a character-level model is more flexible in assigning its capacity to different language pairs. Furthermore, we observe {\nrr that} our multilingual character-level translation even exceeds the quality of bilingual translation in three out of four language pairs, both in BLEU score metric and human evaluation. This demonstrates excellent parameter efficiency of character-level translation in a multilingual setting. We also showcase our model's ability to handle intra-sentence code-switching while performing language identification on the fly.

The contributions of this work are twofold: we empirically show that (1) we can train character-to-character NMT model without any explicit segmentation; and (2) we can share a single character-level encoder across multiple languages to build a multilingual translation system without increasing the model size.


\section{Background: Attentional Neural Machine Translation}\label{sec:back}

Neural machine translation (NMT) is a recently proposed approach to machine translation that builds a single neural network which takes as an input a source sentence $X = (x_1, \ldots, x_{T_X})$ and generates its translation $Y = (y_1, \ldots, y_{T_Y})$, where $x_t$ and $y_{t'}$ are source and target symbols~\cite{Bahdanau:15,Sutskever:14,Luong:15,Cho:14a}. Attentional NMT models have three components: an \ti{encoder}, a \ti{decoder} and an \ti{attention} mechanism. \\

\noindent\textbf{Encoder} Given a source sentence $X$, the encoder constructs a continuous representation that summarizes its meaning with a recurrent neural network (RNN). A bidirectional RNN is often implemented as proposed in \cite{Bahdanau:15}. A forward encoder reads the input sentence from left to right: $\rv{\mb{h}}_t = \fenc\big(\srcemb(x_t), \rv{\mb{h}}_{t-1}\big).$ Similarly, a backward encoder reads it from right to left: $\lv{\mb{h}}_t = \benc\big(\srcemb(x_t), \lv{\mb{h}}_{t+1}\big),$ where $E_x$ is the source embedding lookup table, and $\fenc$ and $\benc$ are recurrent activation functions such as long short-term memory units (LSTMs,~\cite{Hochreiter:97}) or gated recurrent units (GRUs,~\cite{Cho:14b}). The encoder constructs a set of continuous source sentence representations $C$ by concatenating the forward and backward hidden states at each timestep: $C=\big\{\mb{h}_1,\dots,\mb{h}_{T_X}\big\}$, where $\mb{h}_t=\big[\rv{\mb{h}}_t;\lv{\mb{h}}_t\big]$. \\

\noindent\textbf{Attention} First introduced in \cite{Bahdanau:15}, the attention mechanism lets the decoder \ti{attend} more to different source symbols for each target symbol. More concretely, it computes the context vector $\mb{c}_{t'}$ at each decoding time step $t'$ as a weighted sum of the source hidden states: $\mb{c}_{t'} = \sum_{t=1}^{T_X}{\alpha_{{t'}t}\mb{h}_t}.$ {\nrr Similarly to \cite{Chung:16,Firat:16}}, each attentional weight $\alpha_{{t'}t}$ represents how relevant the $t$-th source token $x_t$ is to the $t'$-th target token $y_{t'}$, and is computed as: 
\begin{equation}
\alpha_{t't}=\frac {1} {Z} \fexp\bigg(\fscore\Big(\trgemb(y_{t'-1}),\mb{s}_{t'-1},\mb{h}_{t}\Big)\bigg),
\end{equation}
where $Z={\sum_{k=1}^{T_X}{\fexp\big(\fscore(\trgemb(y_{t'-1}),\mb{s}_{t'-1},\mb{h}_{k})\big)}}$ is the normalization constant. $\fscore()$ is a feedforward neural network with a single hidden layer that scores how well the source symbol $x_t$ and the target symbol $y_{t'}$ match. $\trgemb$ is the target embedding lookup table and $\mb{s}_{t'}$ is the target hidden state at time $t'$. \\

\noindent\tb{Decoder} Given a source context vector $\mb{c}_{t'}$, the decoder computes its hidden state at time $t'$ as: $\mb{s}_{t'} = \dec\big(\trgemb(y_{t'-1}), \mb{s}_{t'-1}, \mb{c}_{t'}\big).$ Then, a parametric function $\foutkp$ returns the conditional probability of the next target symbol being $k$: 
\begin{equation}
\begin{split}
 p(y_{t'}= & k|y_{<t'},X) = \\ 
& \frac{1}{Z}\fexp\bigg(\foutk\Big(\trgemb(y_{t'-1}),\mb{s}_{t'},\mb{c}_{t'}\Big)\bigg)
\end{split}
\end{equation}

where $Z$ is again the normalization constant: $Z={\sum_{j}^{}\fexp\big(\foutj(\trgemb(y_{t'-1}),\mb{s}_{t'},\mb{c}_{t'})\big)}.$ \\

\noindent\tb{Training} The entire model can be trained end-to-end by minimizing the negative conditional log-likelihood, which is defined as:
$$\mathcal{L} = -\frac{1}{N}\sum_{n=1}^{N}\sum_{t=1}^{T_Y^{(n)}}\log p(y_t=y_t^{(n)} | y_{<t}^{(n)},X^{(n)}),$$
where $N$ is the number of sentence pairs, and $X^{(n)}$ and $y_t^{(n)}$ are the source sentence and the $t$-th target symbol in the $n$-th pair, respectively. 


\section{Fully Character-Level Translation}\label{sec:why}

    \subsection{Why Character-Level?}
    
    The benefits of character-level translation over word-level translation are well known. \newcite{Chung:16} present three main arguments: character level models (1) do not suffer from out-of-vocabulary issues, (2) are able to model different, rare morphological variants of a word, and (3) do not require segmentation. Particularly, text segmentation is highly non-trivial for many languages and problematic even for English as word tokenizers are either manually designed or trained on a corpus using an objective function that is unrelated to the translation task at hand, which makes the overall system sub-optimal.
    
    Here we present two additional arguments for character-level translation. First, a character-level translation system can easily be applied to a multilingual translation setting. Between European languages where the majority of alphabets overlaps, for instance, a character-level model may easily identify morphemes that are shared across different languages. A word-level model, however, will need a separate word vocabulary for each language, allowing no cross-lingual parameter sharing.

    Also, by not segmenting source sentences into words, we no longer inject our knowledge of words and word boundaries into the system; instead, we encourage the model to discover an internal structure of a sentence by itself and learn how a sequence of symbols can be mapped to a continuous meaning representation.

    \subsection{Related Work}

    To address these limitations associated with word-level translation, a recent line of research has investigated using sub-word information.
    
    \newcite{Costa-jussa:16} replaced the word-lookup table with convolutional and highway layers on top of character embeddings, while still segmenting source sentences into words. Target sentences were also segmented into words, and prediction was made at word-level.

    Similarly, \newcite{Ling:15} employed a bidirectional LSTM to compose character embeddings into word embeddings. At the target side, another LSTM takes the hidden state of the decoder and generates the target word, character by character. While this system is completely open-vocabulary, it also requires offline segmentation. Also, character-to-word and word-to-character LSTMs significantly slow down training.
    
    Most recently, \newcite{Luong:16} proposed a hybrid scheme that consults character-level information whenever the model encounters an out-of-vocabulary word. As a baseline, they also implemented a purely character-level NMT model with 4 layers of unidirectional LSTMs with 512 cells, with attention over each character. Despite being extremely slow (approximately 3 months to train), the character-level model gave comparable performance to the word-level baseline. This shows the possibility of fully character-level translation.

    Having a word-level decoder restricts the model to only being able to generate previously seen words. \newcite{Sennrich:15} introduced a subword-level NMT model that is capable of open-vocabulary translation using subword-level segmentation based on the byte pair encoding (BPE) algorithm. Starting from a character vocabulary, the algorithm identifies frequent character n-grams in the training data and iteratively adds them to the vocabulary, ultimately giving a subword vocabulary which consists of words, subwords and characters. Once the segmentation rules have been learned, their model performs subword-to-subword translation (\tb{bpe2bpe}) in the same way as word-to-word translation.

    Perhaps the work that is closest to our end goal is \cite{Chung:16}, which used a subword-level encoder from \cite{Sennrich:15} and a fully character-level decoder (\tb{bpe2char}). Their results show that character-level decoding performs better than subword-level decoding. Motivated by this work, we aim for fully character-level translation at both sides (\tb{char2char}).

    Outside NMT, our work is based on a few existing approaches that applied convolutional networks to text, most notably in text classification~\cite{Zhang:15,Xiao:16}. Also, we drew inspiration for our multilingual models from previous work that showed the possibility of training a single recurrent model for multiple languages in domains other than translation~\cite{Tsvetkov:16,Gillick:15}.
    
    \subsection{Challenges}
    
    Sentences are on average 6 (DE, CS and RU) to 8 (FI) times longer when represented in characters. This poses three major challenges to achieving fully character-level translation.
    
    \paragraph{(1) Training/decoding latency} For the decoder, although the sequence to be generated is much longer, each character-level softmax operation costs considerably less compared to a word- or subword-level softmax. \newcite{Chung:16} report that character-level decoding is only 14\% slower than subword-level decoding. 

    On the other hand, computational complexity of the attention mechanism grows quadratically with respect to the sentence length, as it needs to attend to every source token for every target token. This makes a naive character-level approach, such as in \cite{Luong:16}, computationally prohibitive. Consequently, reducing the length of the source sequence is key to ensuring reasonable speed in both training and decoding. 
    
    \paragraph{(2) Mapping character sequence to continuous representation} The arbitrary relationship between the orthography of a word and its meaning is a well-known problem in linguistics~\cite{deSaussure:1916}. Building a character-level encoder is arguably a more difficult problem, as the encoder needs to learn a highly non-linear function from a long sequence of character symbols to a meaning representation. 
    
    \paragraph{(3) Long range dependencies in characters} A character-level encoder needs to model dependencies over longer timespans than a word-level encoder does.
    

\section{Fully Character-Level NMT}\label{sec:model}

    \begin{figure*}[t]
    \centering
    \includegraphics[width=13cm]{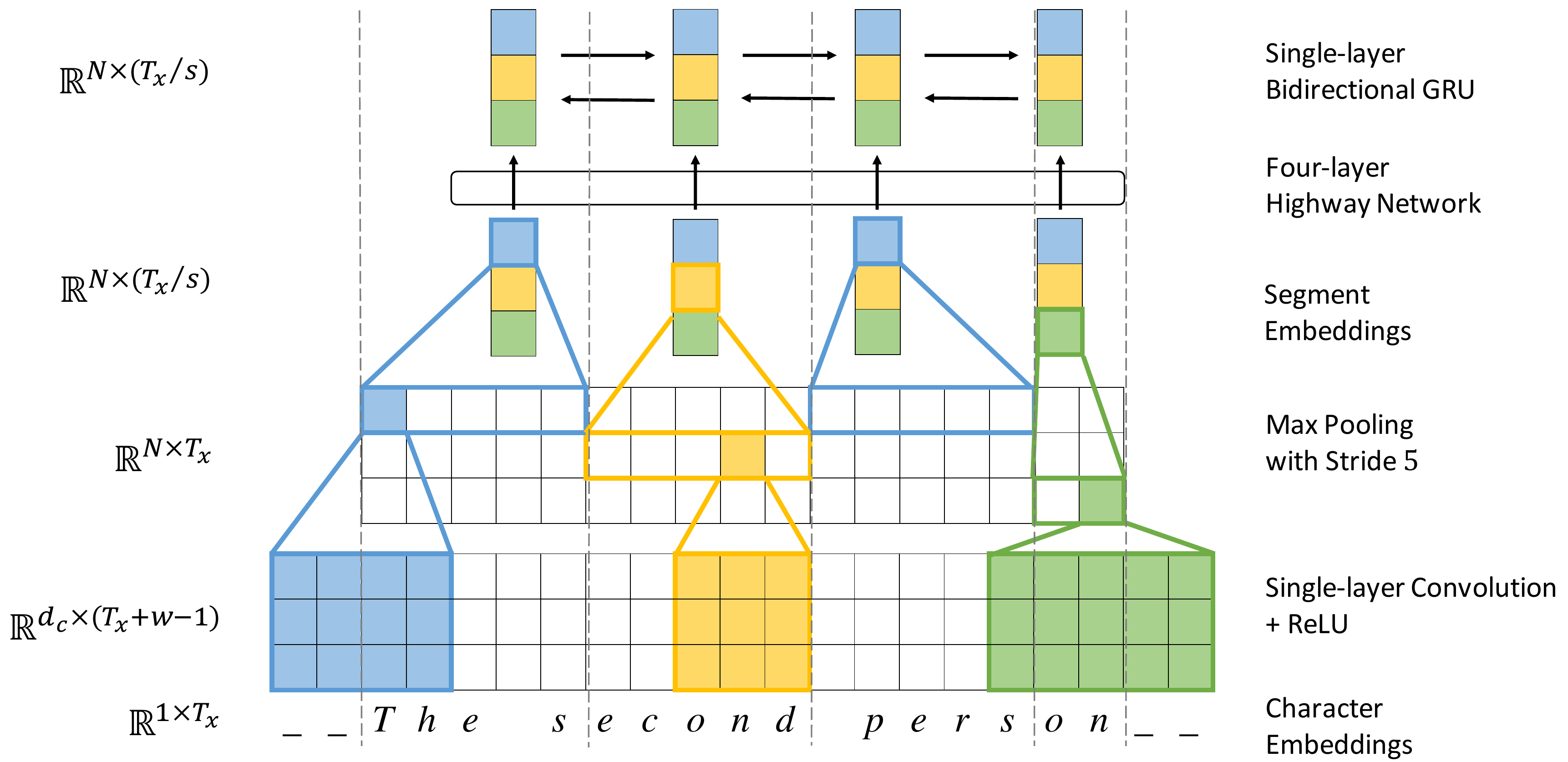}
    \caption{Encoder architecture schematics. Underscore denotes padding. A dotted vertical line delimits each segment. {\nrr The stride of pooling $s$ is 5 in the diagram.} }
    \label{figure:encoder}
    \end{figure*}

    \subsection{Encoder}

    We design an encoder that addresses all the challenges discussed above by using convolutional and pooling layers aggressively to both (1) drastically shorten the input sentence and (2) efficiently capture local regularities. Inspired by the character-level language model from \cite{Kim:15}, our encoder first reduces the source sentence length with a series of convolutional, pooling and highway layers. The shorter representation, instead of the full character sequence, is passed through a bidirectional GRU to (3) help it resolve long term dependencies. We illustrate the proposed encoder in Figure~\ref{figure:encoder} and discuss each layer in detail below. \\
    
    \noindent\textbf{Embedding} We map the {\nrr sequence of source characters $(x_1, \ldots, x_{T_x})$ to a sequence of character embeddings of dimensionality $d_c$: $X = (\mb{C}(x_1), \ldots, \mb{C}(x_{T_x})) \in \mathbb{R}^{d_c \times T_x}$ where $T_x$ is the number of source characters }and $\mb{C}$ is the character embedding lookup table: $\mb{C} \in \mathbb{R}^{d_c \times |\mb{C}|}$.\\
    
    \noindent\textbf{Convolution} One-dimensional convolution operation is then used along consecutive character embeddings. Assuming we have a single filter $\mb{f} \in \mathbb{R}^{d_c \times w}$ of width $w$, we first apply padding to the beginning and the end of $X$, such that the padded sentence $X' \in \mathbb{R}^{d_c \times (T_x + w - 1)}$ is $w-1$ symbols longer. We then apply narrow convolution between $X'$ and $\mb{f}$ such that the $k$-th element of the output $Y_k$ is given as:
    
    \vspace{-4mm}
    \begin{equation}
        Y_k = (X' * \mb{f})_k = \sum_{i,j}^{}{(X'_{[:,k-w+1:k]} \otimes \mb{f})}_{ij},
    \end{equation}
    \vspace{-4mm}

    where $\otimes$ denotes elementwise matrix multiplication and $*$ is the convolution operation. $X'_{[:,k-w+1:k]}$ is the sliced subset of $X'$ that contains all the rows but only $w$ adjacent columns. The padding scheme employed above, commonly known as \ti{half convolution}, ensures the length of the output is identical to the input's: $Y \in \mathbb{R}^{1 \times T_x}.$

    We just illustrated how a single convolutional filter of fixed width might be applied to a sentence. In order to extract informative character patterns of different lengths, we employ a set of filters of varying widths. More concretely, we use a filter bank $\mb{F} = \{\mb{f}_1, \ldots, \mb{f}_{m}\}$ where $\mb{f}_i = \mathbb{R}^{d_c \times i \times n_i}$ is a collection of $n_i$ filters of width $i$. Our model uses $m=8$, hence extracts character n-grams up to $8$ characters long. Outputs from all the filters are stacked upon each other, giving a single representation $Y \in \mathbb{R}^{N \times T_x}$, where the dimensionality of each column is given by the total number of filters $N = \sum_{i=1}^{m}n_i$. Finally, rectified linear activation (ReLU) is applied elementwise to this representation. \\

    \noindent\textbf{Max pooling with stride} The output from the convolutional layer is first split into segments of width $s$, and max-pooling over time is applied to each segment with no overlap. This procedure selects the most salient features to give a \ti{segment embedding}. {\nrr Each segment embedding is a summary of meaningful character n-grams occurring in a particular (overlapping) subsequence in the source sentence. Note that the rightmost segment (above `on') in Figure~\ref{figure:encoder} may capture `son' (the filter in green) although `s' occurs in the previous segment. In other words, our segments are overlapping as opposed to in word- or subword-level models with hard segmentation.}

 Segments act as our internal linguistic unit from this layer and above: the attention mechanism, for instance, attends to each source segment instead of source character. This shortens the source representation $s$-fold: $Y' \in \mathbb{R}^{N \times (T_x / s)}.$ Empirically, we found using smaller $s$ leads to better performance at increased training time. We chose $s=5$ in our experiments as it gives a reasonable balance between the two. \\
    
    \noindent\textbf{Highway network} A sequence of segment embeddings from the max pooling layer is fed into a highway network~\cite{Srivastava:15}. Highway networks are shown to significantly improve the quality of a character-level language model when used with convolutional layers~\cite{Kim:15}. A highway network transforms input $\mb{x}$ with a gating mechanism that adaptively regulates information flow: $$\mb{y} = g \odot \text{ReLU}(\mb{W}_1x+\mb{b}_1) + (1-g) \odot \mb{x},$$ where $g=\sigma((\mb{W}_2\mb{x}+\mb{b}_2)).$ We apply this to each segment embedding individually. \\

    \begin{table}[t]
    \footnotesize
    \begin{center}
    \begin{tabular}{ c | c | c  }  
        \mcn{1}{c}{\tb{Bilingual}} & \mcn{1}{c}{bpe2char} & \mcn{1}{c}{char2char} \\ \hline \hline
        Vocab size & 24,440 & 300 \\ \hline
        Source emb. & 512 & 128 \\ \hline
        Target emb. & 512 & 512 \\ \hline
        \specialcell{Conv.\\filters} &  & \specialcell{200-200-250-250\\-300-300-300-300} \\ \hline
        Pool stride & & 5 \\ \hline
        Highway & & 4 layers \\ \hline
        Encoder & \mcn{2}{c}{1-layer 512 GRUs} \\ \hline
        Decoder & \mcn{2}{c}{2-layer 1024 GRUs} \\ \hline
    \end{tabular}
    \caption{Bilingual model architectures. The char2char model uses 200 filters of width 1, 200 filters of width 2, $\cdots$ and 300 filters of width 8.}
    \vspace{-1mm}
    \label{table:bi-c2c-arch}
    \end{center}
    \normalsize
    \end{table}

    \noindent\textbf{Recurrent layer} Finally, the output from the highway layer is given to a bidirectional GRU from \S\ref{sec:back}, using each segment embedding as input.\\

    {\nrr \noindent\textbf{Subword-level encoder} Unlike a subword-level encoder, our model does not commit to a specific choice of segmentation; it is instead trained to consider every possible character pattern and extract only the most meaningful ones. Therefore, the definition of segmentation in our model is dynamic unlike subword-level encoders. During training, the model finds the most salient character patterns in a sentence via max-pooling, and the character sequences extracted by the model change over the course of training. This is in contrast to how BPE segmentation rules are learned: the segmentation is learned and fixed before training begins.} \\

    \subsection{Attention and Decoder}

    Similarly to the attention model in \cite{Chung:16,Firat:16}, a single-layer feedforward network computes the attention score of next target character to be generated with every source segment representation. {\nrr A standard two-layer character-level decoder then takes the source context vector from the attention mechanism and predicts each target character.} This decoder was described as \textit{base decoder} by \newcite{Chung:16}.

\section{Experiment Settings}\label{sec:exp}

    \subsection{Task and Models}
    
    We evaluate the proposed character-to-character (\tb{char2char}) translation model against subword-level baselines (\tb{bpe2bpe} and \tb{bpe2char}) on the WMT'15 DE$\ra$EN, CS$\ra$EN, FI$\ra$EN and RU$\ra$EN translation tasks.\footnote{\tt\href{http://www.statmt.org/wmt15/translation-task.html}{http://www.statmt.org/wmt15/translation\\-task.html}} We do not consider word-level models, as it has already been shown that subword-level models outperform them by mitigating issues inherent to closed-vocabulary translation~\cite{Sennrich:15,Sennrich:16}. Indeed, subword-level NMT models have been the de-facto state-of-the-art and are now used in a very large-scale industry NMT system to serve millions of users per day~\cite{Wu:16}.
    
    We experiment in two different scenarios: 1) a bilingual setting where we train a model on data from a single language pair; and 2) a multilingual setting where the task is many-to-one translation: we train a single model on data from all four language pairs. Hence, our baselines and models are: 
    
    \begin{enumerate}[label=(\alph*)]
    \itemsep-0.3em 
    \item bilingual bpe2bpe: from~\cite{Firat:16}.
    \item bilingual bpe2char: from~\cite{Chung:16}.
    \item bilingual char2char
    \item multilingual bpe2char
    \item multilingual char2char
    \end{enumerate}

    \noindent We train all the models ourselves other than (a), for which we report the results from \cite{Firat:16}. We detail the configuration of our models in Table~\ref{table:bi-c2c-arch} and Table~\ref{table:bi-c2c-multi-arch}.

    \subsection{Datasets and Preprocessing}
    
    We use all available parallel data on the four language pairs from WMT'15: DE-EN, CS-EN, FI-EN and RU-EN.
    
    For the bpe2char baselines, we only use sentence pairs where the source is no longer than 50 subword symbols. For our char2char models, we only use pairs where the source sentence is no longer than 450 characters. For all the language pairs apart from FI-EN, we use newstest-2013 as a development set and newstest-2014 and newstest-2015 as test sets. For FI-EN, we use newsdev-2015 and newstest-2015 as development and test sets respectively. We tokenize\footnote{This is unnecessary for char2char models, yet was carried out for comparison.} each corpus using the script from Moses.\footnote{\tt\href{https://github.com/moses-smt/mosesdecoder}{https://github.com/moses-smt/mosesdecod\\er}}

    When training bilingual bpe2char models, we extract 20,000 BPE operations from each of the source and target corpus using a script from \cite{Sennrich:15}. This gives a source BPE vocabulary of size 20k$-$24k for each language.

    \begin{table}[t]
    \footnotesize
    \begin{center}
    \begin{tabular}{ c | c | c  }  
        \mcn{1}{c}{\tb{Multilingual}} & \mcn{1}{c}{bpe2char} & \mcn{1}{c}{char2char} \\ \hline \hline
        Vocab size & {54,544} & {400} \\ \hline
        Source emb. & 512 & 128 \\ \hline
        Target emb. & 512 & 512 \\ \hline
        \specialcell{Conv.\\filters} &  & \specialcell{{200-250-300-300}\\{-400-400-400-400}} \\ \hline
        Pool stride & & 5 \\ \hline
        Highway & & 4 layers \\ \hline
        Encoder & \mcn{2}{c}{1-layer 512 GRUs} \\ \hline
        Decoder & \mcn{2}{c}{2-layer 1024 GRUs} \\ \hline
    \end{tabular}
    \caption{Multilingual model architectures.}
    \vspace{-2mm}
    \label{table:bi-c2c-multi-arch}
    \end{center}
    \normalsize
    \end{table}

    \subsection{Training Details}

    Each model is trained using stochastic gradient descent and Adam~\cite{Kingma:14} with learning rate $0.0001$ and minibatch size 64. Training continues until the BLEU score on the validation set stops improving. The norm of the gradient is clipped with a threshold of 1~\cite{Pascanu:13}. All weights are initialized from a uniform distribution $[-0.01, 0.01].$ 

    Each model is trained on a single pre-2016 GTX Titan X GPU with 12GB RAM.

    \subsection{Decoding Details}
    
    As from \cite{Chung:16}, a two-layer unidirectional character-level decoder with 1024 GRU units is used for all our experiments. For decoding, we use beam search with length-normalization to penalize shorter hypotheses. The beam width is 20 for all models.
    
    \subsection{Training Multilingual Models}

    \noindent\tb{Task description} We train a model on a many-to-one translation task to translate a sentence in any of the four languages (German, Czech, Finnish and Russian) to English. We do \ti{not} provide a language identifier to the encoder, but merely the sentence itself, encouraging the model to perform language identification on the fly. In addition, by not providing the language identifier, we expect the model to handle intra-sentence code-switching seamlessly. \\

    \noindent\tb{Model architecture} The multilingual char2char model uses slightly more convolutional filters than the bilingual char2char model, namely (200-250-300-300-400-400-400-400). Otherwise, the architecture remains the same as shown in Table~\ref{table:bi-c2c-arch}. By not changing the size of the encoder and the decoder, we fix the capacity of the core translation module, and only allow the multilingual model to detect more character patterns.
    
    Similarly, the multilingual bpe2char model has the same encoder and decoder as the bilingual bpe2char model, but a larger vocabulary. We learn 50,000 multilingual BPE operations on the multilingual corpus, resulting in 54,544 subwords. See Table~\ref{table:bi-c2c-multi-arch} for the exact configuration of our multilingual models. \\

    \noindent\tb{Data scheduling} For the multilingual models, an appropriate scheduling of data from different languages is crucial to avoid overfitting to one language too soon. Following \cite{Firat:16,Firat:16b}, each minibatch is \ti{balanced}, in that the proportion of each language pair in a single minibatch corresponds to that of the full corpus. With this minibatch scheme, roughly the same number of updates is required to make one full pass over the entire training corpus of each language pair. Minibatches from all language pairs are combined and presented to the model as a single minibatch. See Table~\ref{table:corpus-stats} for the minibatch size for each language pair.

        \begin{table}[h]
        \footnotesize
        \begin{center}
        \begin{tabular}{ r | r | r | r | r } 
            \multicolumn{1}{c}{} & \multicolumn{1}{r}{DE-EN} & \multicolumn{1}{r}{CS-EN} & \multicolumn{1}{r}{FI-EN} & \multicolumn{1}{r}{RU-EN} \\ \hline \hline
            corpus size    & 4.5m & 12.1m & 1.9m & 2.3m \\
            minibatch size & 14 & 37 & 6 & 7 \\ \hline
        \end{tabular}
        \caption{The minibatch size of each language (second row) is proportionate to the number of sentence pairs in each corpus (first row).}
        \label{table:corpus-stats}
        \end{center}
        \normalsize
        \end{table}

    \noindent\tb{Treatment of Cyrillic} To facilitate cross-lingual parameter sharing, we convert every Cyrillic character in the Russian source corpus to Latin alphabet according to ISO-9. Table~\ref{table:iso9} shows an example of how this conversion may help the multilingual models identify lexemes that are shared across multiple languages.

        \begin{table}[h]
        \vspace{-2mm}
        \footnotesize
        \begin{center}
        \begin{tabular}{ l | r | r  } 
            \multicolumn{1}{c}{} & \multicolumn{1}{r}{school} & \multicolumn{1}{r}{schools} \\ \hline \hline
            CS & \foreignlanguage{czech}{škola} & \foreignlanguage{czech}{školy} \\
            RU & \foreignlanguage{russian}{школа} & \foreignlanguage{russian}{школы} \\
            RU (ISO-9) & \foreignlanguage{czech}{škola} & \foreignlanguage{czech}{školy} \\ \hline
        \end{tabular}
        \caption{Czech and Russian words for \ti{school} and \ti{schools}, alongside the conversion of Russian characters into Latin.}
        \vspace{-2mm}
        \label{table:iso9}
        \end{center}
        \normalsize
        \end{table}

    \begin{table*}[ht!]
        \footnotesize
        \centering
        \begin{tabular}{*{55}{r|l|r|l|l||K{1.5cm}|K{1.5cm}|K{1.5cm}}}
        
                               \mcn{3}{r}{Setting} &  \mcn{1}{c}{Src} & \mcn{1}{c||}{Trg}   & \mcn{1}{c}{Dev} & \mcn{1}{c}{Test1} & \mcn{1}{c}{Test2} \\
                               \hline \hline
                               
        \multirow{6}{*}{DE-EN} & $\text{(a)}^{*}$ & bi & bpe & bpe & 24.13    &       & 24.00      \\
                               & (b) & bi & bpe & char & 25.64    & 24.59      & 25.27      \\
                               & (c) & bi & char & char & \tb{26.30}    & \tb{25.77}      & \tb{25.83}      \\ \cline{2-8}

                               & (d) & multi & bpe & char & 24.92    & 24.54      & 25.23      \\
                               & (e) & multi & char & char & 25.67    & 25.13      & 25.79      \\ \hline \hline
                               
        \multirow{6}{*}{CS-EN} & $\text{(f)}^{*}$ & bi & bpe & bpe & 21.24     &       & 20.32      \\
                               & (g) & bi & bpe & char & 22.95     & 23.78      & 22.40      \\
                               & (h) & bi & char & char & 23.38     & 24.08      & 22.46     \\ \cline{2-8}

                               & (i) & multi & bpe & char & 23.27     & 24.27      & 22.42      \\
                               & (j) & multi & char & char & \tb{24.09}     & \tb{25.01}      & \tb{23.24}      \\\hline \hline
                               
        \multirow{6}{*}{FI-EN} & $\text{(k)}^{*}$ & bi & bpe & bpe & 13.15     &       & 12.24      \\
                               & (l) & bi & bpe & char & 14.54     &       & 13.98      \\
                               & (m) & bi & char & char & 14.18     &       & 13.10      \\ \cline{2-8}

                               & (n) & multi & bpe & char & 14.70     &       & 14.40      \\
                               & (o) & multi & char & char & \tb{15.96}     &       & \tb{15.74}      \\ \hline \hline
                               
        \multirow{6}{*}{RU-EN} & $\text{(p)}^{*}$ & bi & bpe & bpe & 21.04     &       & 22.44  \\
                               & (q) & bi & bpe & char & 21.68     & 26.21      & {22.83} \\
                               & (r) & bi & char & char & 21.75     & \tb{26.80}      & 22.73      \\ \cline{2-8}

                               & (s) & multi & bpe & char & 21.75     & 26.31      & 22.81      \\
                               & (t) & multi & char & char & \tb{22.20}     & 26.33      & \tb{23.33} \\ \hline
        \end{tabular}
        \caption{BLEU scores of five different models on four language pairs. For each test or development set, the best performing model is shown in bold. $(*)$ results are taken from \protect\cite{Firat:16}. }
        \label{table:results}
        \normalsize
    \end{table*}

    \noindent\tb{Multilingual BPE} For the multilingual bpe2char model, multilingual BPE segmentation rules are extracted from a large dataset containing training source corpora of all the language pairs. To ensure the BPE rules are not biased towards one language, larger datasets such as Czech and German corpora are trimmed such that every corpus contains an approximately equal number of characters. \\


\vspace{-5mm}
\section{Quantitative Analysis}\label{sec:quant}

    \begin{table*}[ht!]
        \footnotesize
        \centering
        \begin{tabular}{*{55}{r|l|r|l|l||K{1.0cm}|K{1.0cm}||K{1.0cm}|K{1.0cm}}}
        
								\mcn{5}{c||}{} & \mcn{2}{c||}{Adequacy} & \mcn{2}{c}{Fluency} \\ \cline{6-9}
                               \mcn{3}{r}{Setting} &  \mcn{1}{c}{Src} & \mcn{1}{c||}{Trg}   & \mcn{1}{c}{Raw (\%)} & \mcn{1}{c||}{Stnd. ($\sigma$)} & \mcn{1}{c}{Raw (\%)} & \mcn{1}{c}{Stnd. ($\sigma$)} \\
                               \hline \hline
                               
        \multirow{4}{*}{DE-EN} & (a) & bi & bpe & char & 65.47 & -0.0536 & 68.64 & 0.0052         \\
                               & (b) & bi & char & char & 68.11  & \tb{0.0509} & 68.80 & 0.0468    \\ 
                               & (c) & multi & char & char & 67.80 & \tb{0.0281} & 68.92 & 0.0282            \\ \hline \hline
                               
        \multirow{4}{*}{CS-EN} & (d) & bi & bpe & char & 62.76 & \tb{0.0361} & 61.62 & -0.0285       \\
                               & (e) & bi & char & char & 60.78 & -0.0154 & 63.37 & 0.0410            \\ 
                               & (f) & multi & char & char & 63.03 & \tb{0.0415} & 65.08 & \tb{0.1047}         \\\hline \hline
                               
        \multirow{4}{*}{FI-EN} & (g) & bi & bpe & char & 47.03 & -0.1326 & 59.33 & -0.0329       \\
                               & (h) & bi & char & char & 50.17 & \tb{-0.0650} & 59.97 & -0.0216      \\ 
                               & (i) & multi & char & char & 50.95 & \tb{-0.0110} & 63.26 & \tb{0.0969}           \\ \hline \hline
                               
        \multirow{4}{*}{RU-EN} & (j) & bi & bpe & char & 61.26 & -0.1062 & 57.74 & -0.0592   \\
                               & (k) & bi & char & char & 64.06 & \tb{0.0105} & 59.85 & 0.0168             \\ 
                               & (l) & multi & char & char & 64.77 & \tb{0.0116} & 63.32 & \tb{0.1748}        \\ \hline
        \end{tabular}
        \caption{Human evaluation results for adequacy and fluency. We present both the averaged raw scores (Raw) and the averaged standardized scores (Stnd.). Standardized adequacy is used to rank the systems and standardized fluency is used to break ties. {\nrr A positive standardized score should be interpreted as the number of standard deviations above this particular worker's mean score that this system scored on average.} For each language pair, we boldface the best performing model with statistical significance. When there is a tie, we boldface both systems. }
        \label{table:human}
        \normalsize
    \end{table*}

    \subsection{Evaluation with BLEU Score}

    In this section, we first establish our main hypotheses for introducing character-level and multilingual models, and investigate whether our observations support or disagree with our hypotheses. From our empirical results, we want to verify: (1) if fully character-level translation outperforms subword-level translation, (2) in which setting and to what extent is multilingual translation beneficial and (3) if multilingual, character-level translation achieves superior performance to other models. We outline our results with respect to each hypothesis below. \\

    \noindent\tb{(1) Character- vs. subword-level} In a bilingual setting, the char2char model outperforms both subword-level baselines on DE-EN (Table~\ref{table:results}~(a-c)) and CS-EN (Table~\ref{table:results}~(f-h)). On the other two language pairs, it exceeds the bpe2bpe model and achieves similar performance with the bpe2char baseline (Table~\ref{table:results}~(k-m) and (p-r)). We conclude that the proposed character-level model is comparable to or better than both subword-level baselines.

    Meanwhile, in a multilingual setting, the character-level encoder significantly surpasses the subword-level encoder consistently in all the language pairs (Table~\ref{table:results}~(d-e), (i-j), (n-o) and (s-t)). From this, we conclude that translating at the level of characters allows the model to discover shared constructs between languages more effectively. This also demonstrates that the character-level model is more flexible in assigning model capacity to different language pairs.\\

    \noindent\tb{(2) Multilingual vs. bilingual} At the level of characters, we note that multilingual translation is indeed strongly beneficial. On the test sets, the multilingual character-level model outperforms the single-pair character-level model by 2.64 BLEU in FI-EN (Table~\ref{table:results}~(m, o)) and 0.78 BLEU in CS-EN (Table~\ref{table:results}~(h, j)), while achieving comparable results on DE-EN and RU-EN. 

    At the level of subwords, on the other hand, we do not observe the same degree of performance benefit from multilingual translation. Also, the multilingual bpe2char model requires much more updates to reach the performance of the bilingual bpe2char model (see Figure~\ref{figure:overfit}). This suggests that learning useful subword segmentation across languages is difficult. \\

    \noindent\tb{(3) Multilingual char2char vs. others} The multilingual char2char model is the best performer in CS-EN, FI-EN and RU-EN (Table~\ref{table:results}~(j, o, t)), and is the runner-up in DE-EN (Table~\ref{table:results}~(e)). The fact that the multilingual char2char model outperforms the single-pair models goes to show the parameter efficiency of character-level translation: instead of training $N$ separate models for $N$ language pairs, it is possible to get better performance with a single multilingual character-level model.

    \begin{table*}[h!]
    \footnotesize
    \centering

    \begin{tabular}{p{1.4cm}|p{14.2cm}}
    \multicolumn{2}{l}{\tb{(a) Spelling mistakes}} \\ \hline \hline
    DE ori  & Warum sollt{\clr en} wir nich{\clr t} Freunde sei ? \\ \hline
    DE src  & Warum sollt{\clr ne} wir nich Freunde sei ? \\ \hline
    EN ref  & Why {\clr should not we be} friends ? \\ \hline
    bpe2char &  Why {\clr are we to be} friends ?\\ \hline
    char2char & Why {\clr should we not be} friends ? \\ \hline
    \multicolumn{2}{l}{}       \\

    \multicolumn{2}{l}{\tb{(b) Rare words}} \\ \hline \hline
    DE src  & \vspace{-0.80em}{\clr Siebentausend}zweihundertvierundf\"unfzig . \\ \hline
    EN ref  & \vspace{-0.80em}{\clr Seven thousand} two hundred fifty four . \\ \hline
    bpe2char &  Fifty-five {\clr Decline of the Seventy} . \\ \hline
    char2char & \vspace{-0.80em}{\clr Seven thousand} hundred thousand fifties . \\ \hline
    \multicolumn{2}{l}{}       \\

    \multicolumn{2}{l}{\tb{(c) Morphology}} \\ \hline \hline
    DE src  & Die Zufahrtsstraßen wurden {\clr gesperrt} , wodurch sich laut CNN lange R\"uckstaus bildeten . \\ \hline
    EN ref  & The access roads were {\clr blocked off} , which , according to CNN , caused long tailbacks . \\ \hline
    bpe2char &  The access roads were {\clr locked} , which , according to CNN , was long back . \\ \hline
    char2char & The access roads were {\clr blocked} , which looked long backwards , according to CNN . \\ \hline
    \multicolumn{2}{l}{}       \\

    \multicolumn{2}{l}{\tb{(d) Nonce words}} \\ \hline \hline
    DE src  & Der Test ist nun \"uber , aber ich habe keine gute Note . Es ist wie eine {\clr Verschlimmbesserung} . \\ \hline
    EN ref  & The test is now over , but i don't have any good grade . it is like a {\clr worsened improvement} . \\ \hline
    bpe2char &  The test is now over , but i do not have a good note . \\ \hline
    char2char & The test is now , but i have no good note , it is like a {\clr worsening improvement} . \\ \hline
    \multicolumn{2}{l}{}       \\

    \multicolumn{2}{l}{\tb{(e) Multilingual}} \\ \hline \hline
    Multi src  & \vspace{-0.85em} {\clg\foreignlanguage{german}{ Bei der}} {\cly \foreignlanguage{czech}{Metropolitního výboru pro dopravu}} {\clg\foreignlanguage{german}{für das Gebiet der San Francisco Bay erklärten Beamte , der Kongress könne das Problem}} {\clb\foreignlanguage{russian}{банкротство доверительного Фонда строительства шоссейных дорог}} {\clg\foreignlanguage{german}{einfach durch Erhöhung der Kraftstoffsteuer lösen }} .\\ \hline
    EN ref  & At the Metropolitan Transportation Commission in the San Francisco Bay Area , officials say Congress could very simply {\clr deal with the bankrupt Highway Trust Fund }by raising gas taxes . \\ \hline
    bpe2char & During the Metropolitan Committee on Transport for San Francisco Bay , officials declared that Congress could {\clr solve the problem of bankruptcy} by increasing the fuel tax bankrupt . \\ \hline
    char2char & At the Metropolitan Committee on Transport for the territory of San Francisco Bay , officials explained that the Congress could simply {\clr solve the problem of the bankruptcy of the Road Construction Fund }by increasing the fuel tax . \\ \hline
    \multicolumn{2}{l}{}       \\

    \end{tabular}
    \vspace{-8mm}
    \caption{Sample translations. For each example, we show the source sentence as \ti{src}, the human translation as \ti{ref}, and the translations from the subword-level baseline and our character-level model as \ti{bpe2char} and \ti{char2char}, respectively. For (a), the original, uncorrupted source sentence is also shown (\ti{ori}). The source sentence in (e) contains words in German (in green), Czech (in yellow) and Russian (in blue). The translations in (a-d) are from the bilingual models, whereas those in (e) are from the multilingual models.}
    \label{table:qual}
    \normalsize
    \end{table*}

    \subsection{Human Evaluation}

    It {\nrr is} well known that automatic evaluation metrics such as BLEU encourage reference-like translations and do not fully capture true translation quality~\cite{Callison-Burch:09,Graham:15}. Therefore, we also carry out a recently proposed evaluation from \cite{Graham:16} where we have human assessors rate both (1) adequacy and (2) fluency of each system translation on a scale from 0 to 100 via {\nrr Amazon Mechanical Turk}. Adequacy is the degree to which assessors agree that the system translation expresses the meaning of the reference translation. Fluency is evaluated using system translation alone without any reference translation. 
    
    {\nrr Approximately 1k turkers assessed a single test set (3k sentences in newstest-2014) for each system and language pair. Each turker conducted a minimum of 100 assessments for quality control, and the set of scores generated by each turker was standardized to remove any bias in the individual's scoring strategy.}

    We consider three models (bilingual bpe2char, bilingual char2char and multilingual char2char) for the human evaluation. We leave out the multilingual bpe2char model to minimize the number of similar systems to improve the interpretability of the evaluation overall.

    For DE-EN, we observe that the multilingual char2char and bilingual char2char models are tied with respect to both adequacy and fluency (Table~\ref{table:human}~(b-c)). For CS-EN, the multilingual char2char and bilingual bpe2char models ared tied for adequacy. However, the multilingual char2char model yields significantly better fluency (Table~\ref{table:human}~(d, f)). For FI-EN and RU-EN, the multilingual char2char model is tied with the bilingual char2char model with respect to adequacy, but significantly outperforms all other models in fluency (Table~\ref{table:human}~(g-i, j-l)). 
    
    Overall, the improvement in translation quality yielded by the multilingual character-level model mainly comes from fluency. We conjecture that because the English decoder of the multilingual model is tuned on all the training sentence pairs, it becomes a better language model than a bilingual model's decoder. We leave it for future work to confirm if this is indeed the case.

\section{Qualitative Analysis}\label{sec:qual}

In Table~\ref{table:qual}, we demonstrate our character-level model's robustness in four translation scenarios that conventional NMT systems are known to suffer in. We also showcase our model's ability to seamlessly handle intra-sentence \ti{code-switching}, or mixed utterances from two or more languages. We compare sample translations from the character-level model with those from the subword-level model, which already sidesteps some of the issues associated with word-level translation.

With real-world text containing typos and spelling mistakes, the quality of word-based translation would severely drop, as every non-canonical form of a word cannot be represented. On the other hand, a character-level model has a much better chance recovering the original word or sentence. Indeed, our char2char model is robust against a few spelling mistakes (Table~\ref{table:qual}~(a)).

Given a long, rare word such as ``Siebentausendzweihundertvierundf\"unfzig'' (seven thousand two hundred fifty four) in Table~\ref{table:qual}~(b), {\nrr the subword-level model segments ``Siebentausend'' as (Sieb, ent, aus, end), which results in an inaccurate translation.} The character-level model performs better on these long, concatenative words with ambiguous segmentation.

Also, we expect a character-level model to handle novel and unseen morphological inflections well. We observe that this is indeed the case, as our char2char model correctly understands ``gesperrt'', a past participle form of ``sperren'' (to block) (Table~\ref{table:qual}~(c)).

Nonce words are terms coined for a single use. They are not actual words but are constructed in a way that humans can intuitively guess what they mean, such as \ti{workoliday} and \ti{friyay}. We construct a few DE-EN sentence pairs that contain German nonce words (one example shown in Table~\ref{table:qual}~(d)), and observe that the character-level model can indeed detect salient character patterns and arrive at a correct translation.

Finally, we evaluate our multilingual models' capacity to perform intra-sentence code-switching, by giving them as input mixed sentences from multiple languages. {\nrr The newstest-2013 development datasets for DE-EN, CS-EN and FI-EN contain intersecting examples with the same English sentences. We compile a list of these sentences in DE/CS/FI and their translation in EN, and choose a few samples uniformly at random from the English side. Words or clauses from different languages are manually intermixed to create multilingual sentences. }

We discover that when given sentences with high degree of language intermixing, as in Table~\ref{table:qual}~(e), the multilingual bpe2char model fails to seamlessly handle alternation of languages. Overall, however, both multilingual models generate reasonable translations. This is possible because we did not provide a language identifier when training our multilingual models; as a result, they learned to understand a multilingual sentence and translate it into a coherent English sentence. We show supplementary sample translations in each scenario on a webpage.\footnote{\tt\href{https://sites.google.com/site/dl4mtc2c}{https://sites.google.com/site/dl4mtc2c}}

\noindent\tb{Training and decoding speed} On a single Titan X GPU, we observe that our char2char models are approximately 35\% slower to train than our bpe2char baselines when the same batch size was used. Our bilingual character-level models can be trained in roughly two weeks. 

    We further note that the bilingual bpe2char model can translate 3,000 sentences in 66.63 minutes while the bilingual char2char model requires 71.71 minutes (online, not in batch). See Table~\ref{table:model-speed} for the exact details. \\

\begin{table}[h]
\footnotesize
\begin{center}
\vspace{-4mm}
\hspace{-6mm}
\begin{tabular}{ c  c | c c | c }  
    \multicolumn{1}{c}{} & \multicolumn{1}{c|}{Model} & \multicolumn{1}{c}{\specialcell{Time to\\execute 1k\\updates (s)}} & \multicolumn{1}{c|}{\specialcell{Batch\\size}}  & \multicolumn{1}{c}{\specialcell{Time to\\decode 3k\\sentences (m)}} \\ \hhline{~====}
    FI-EN & bpe2char & $2461.72$ & 128 & 66.63 \\ \hhline{~----}
    & char2char & $2371.93$ & 64 & 71.71 \\ \hhline{~====}

    Multi & bpe2char & $1646.37$ & 64 & 68.99  \\ \hhline{~----}
    & char2char & $2514.23$ & 64 & 72.33  \\ \hhline{~----}
\end{tabular}
\caption{Speed comparison. The second column shows the time taken to execute 1,000 training updates. The model makes each update after having seen one minibatch.}
\vspace{-2mm}
\label{table:model-speed}
\end{center}
\normalsize
\end{table}

\begin{figure}[h!]
\centering
\hspace{-6mm}
\includegraphics[width=80mm]{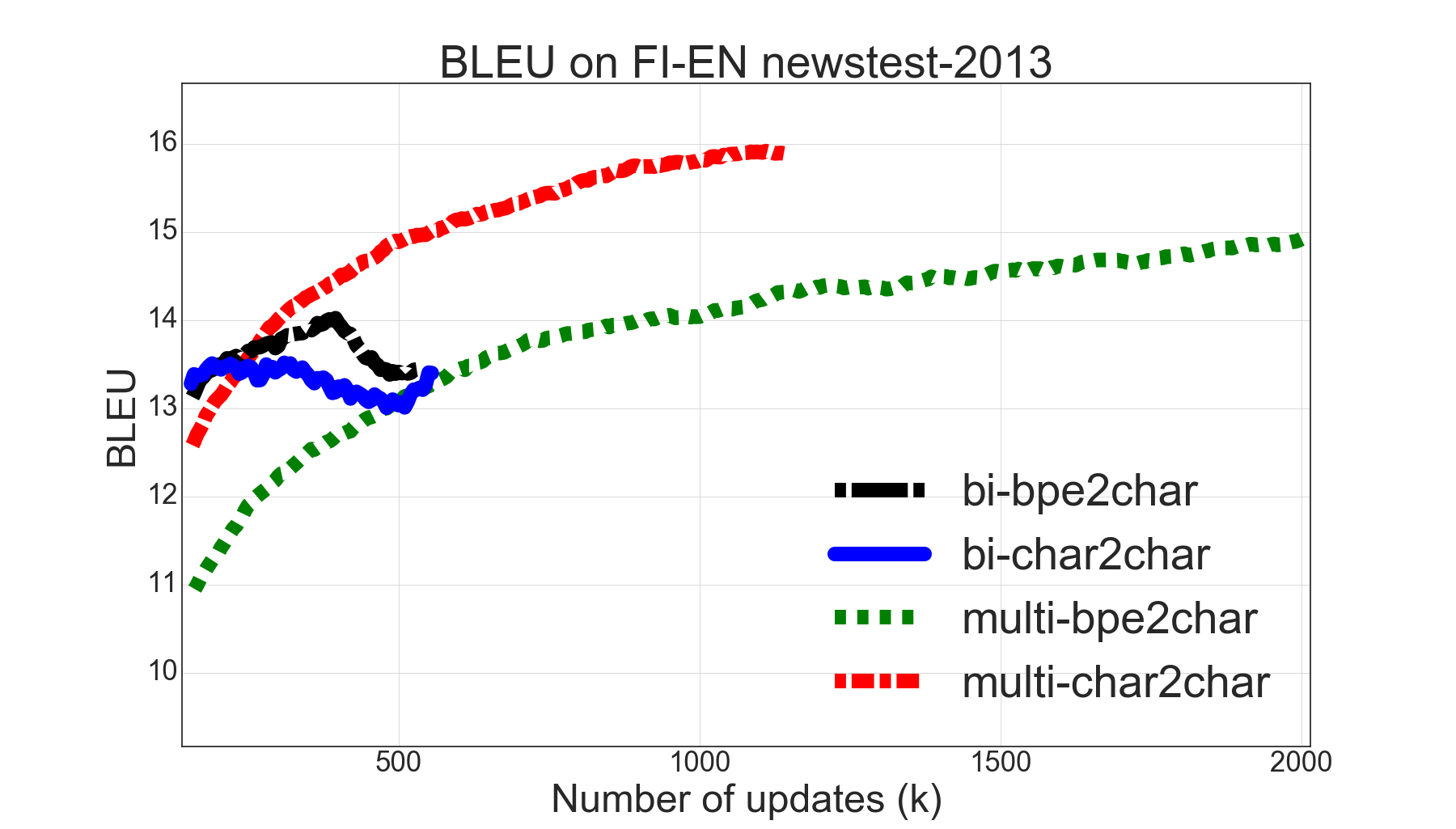}
\caption{Multilingual models overfit less than bilingual models on low-resource language pairs.}
\label{figure:overfit}
\vspace{-4mm}
\end{figure}

\noindent\tb{Further observations} We also note that the multilingual models are less prone to overfitting than the bilingual models. This is particularly visible for low-resource language pairs such as FI-EN. Figure~\ref{figure:overfit} shows the evolution of the FI-EN validation BLEU scores where the bilingual models overfit rapidly but the multilingual models seem to regularize learning by training simultaneously on other language pairs.\\

\section{Conclusion}\label{sec:end}

We propose a fully character-level NMT model that accepts a sequence of characters in the source language and outputs a sequence of characters in the target language. What is remarkable about this model is the absence of explicitly hard-coded knowledge of words and their boundaries, and that the model learns these concepts from a translation task alone.

Our empirical results show that the fully character-level model performs as well as, or better than, subword-level translation models. The performance gain is distinctly pronounced in the multilingual many-to-one translation task, where results show that character-level model can assign model capacities to different languages more efficiently than the subword-level models. We observe a particularly large improvement in FI-EN translation when the model is trained to translate multiple languages, indicating positive cross-lingual transfer to a low-resource language pair.

We discover two main benefits of the multilingual character-level model: (1) it is much more parameter efficient than the bilingual models and (2) it can naturally handle intra-sentence code-switching as a result of the many-to-one translation task. Ultimately, we present a case for fully character-level translation: that translation at the level of character is strongly beneficial and should be encouraged more.

The repository {\tt\small\href{https://github.com/nyu-dl/dl4mt-c2c}{https://github.com/nyu-dl\\/dl4mt-c2c}} contains the source code and pre-trained models for reproducing the experimental results.

In the next stage of this research, we will investigate extending our multilingual many-to-one translation models to perform many-to-many translation, which will allow the decoder, similarly with the encoder, to learn from multiple target languages. Furthermore, a more thorough investigation into model architectures and hyperparameters is needed.

\section*{Acknowledgements}

KC thanks the support by eBay, Facebook, Google (Google Faculty Award 2016) and NVidia (NVIDIA AI Lab 2016-2019). This work was partly supported by Samsung Advanced Institute of Technology (Deep Learning). JL was supported by Qualcomm Innovation Fellowship, and thanks David Yenicelik and Kevin Wallimann for their contribution in designing the qualitative analysis. The authors would like to thank Prof. Zheng Zhang (NYU Shanghai) for fruitful discussion and comments, as well as Yvette Graham for her help with the human evaluation. 

\bibliography{paper}
\bibliographystyle{acl2012}

\/*

\appendix

\section{Supplementary Examples}

We show additional sample translations in five scenarios: spelling mistakes (Table~\ref{table:typo}), rare and long words (Table~\ref{table:rare}), nonce words (Table~\ref{table:nonce}), morphological inflections (Table~\ref{table:morph}) and intra-sentence code-switching (Table~\ref{table:multilingual}).

\begin{table*}[t]
\footnotesize
\centering

\begin{tabular}{p{1.4cm}|p{14.6cm}}

\multicolumn{2}{l}{\tb{(a) Spelling mistakes}} \\ \hline \hline
DE ori  & Die Premi{\clr er}min{\clr i}ster Indiens und Japans trafen s{\clr i}ch in Tokio . \\ \hline
DE src  & Die Premi{\clr re}minster Indiens und Japans trafen sch in Tokio . \\ \hline
EN ref  & India and Japan {\clr prime ministers} meet in Tokyo . \\ \hline
bpe2char &  The {\clr Premier Minster} of India and Japan met in Tokyo . \\ \hline
char2char & The {\clr prime ministers} of India and Japan met in Tokyo . \\ \hline
\multicolumn{2}{l}{}       \\

\multicolumn{2}{l}{\tb{(b) Spelling mistakes}} \\ \hline \hline
DE ori  & Wir haben nichts zu verli{\clr er}en . \\ \hline
DE src  & Wir haben nichts zu verli{\clr rr}en . \\ \hline
EN ref  & We have nothing to {\clr lose} . \\ \hline
bpe2char & We have nothing to {\clr limit} . \\ \hline
char2char & We have nothing to {\clr lose} . \\ \hline
\multicolumn{2}{l}{}       \\

\end{tabular}
\vspace{-8mm}
\caption{Sample translations of German sentences with spelling mistakes.}
\label{table:typo}
\normalsize
\end{table*}

\begin{table*}[t]
\footnotesize
\centering
\begin{tabular}{p{1.4cm}|p{14.6cm}}

\multicolumn{2}{l}{\tb{(c) Rare words}} \\ \hline \hline
DE src  & Die {\clr Kluser-Ampel} sichere sowohl Radfahrer als auch Busfahrg\"aste und die Bergle-Bewohner . \\ \hline
EN ref  & The {\clr Kluser lights} protect cyclists , as well as those travelling by bus and the residents of Bergle . \\ \hline
bpe2char & The {\clr cluster traffic lights} secure both cyclists and bus passengers and mountain residents . \\ \hline
char2char &  The {\clr Kluser traffic lights} secure both cyclists and bus passengers and the mountain mountain residents . \\ \hline
\multicolumn{2}{l}{}       \\

\multicolumn{2}{l}{\tb{(d) Rare words}} \\ \hline \hline
DE src  & In Swasiland leben 245000 {\clr Aids-waisen} . \\ \hline
EN ref  & In Swaziland , there are 245,000 {\clr AIDS orphans} . \\ \hline
bpe2char & In Swaziland , 245,000 {\clr AIds-waisen} live . \\ \hline
char2char & In Swaziland , 245000 {\clr AIDS orphans} live . \\ \hline
\multicolumn{2}{l}{}       \\

\multicolumn{2}{l}{\tb{(e) Long words}} \\ \hline \hline
DE src  & Bezirks{\clr schornsteinfegermeister} . \\ \hline
EN ref  & District {\clr chimney sweep master} .  \\ \hline
bpe2char &  District of the {\clr district of the district} . \\ \hline
char2char & Residential {\clr Chimney Mayors} . \\ \hline
\multicolumn{2}{l}{}       \\

\multicolumn{2}{l}{\tb{(f) Long words}} \\ \hline \hline
DE src  & Ich bin keine {\clr Einhundert-Dollar-Note} , die jedem gef\"allt . \\ \hline
EN ref  & I am not a {\clr hundred dollar bill} to please all . \\ \hline
bpe2char &  I am not a {\clr one-hundred dollar touch} that everyone likes . \\ \hline
char2char & I am not a {\clr hundred dollar notes} that everyone likes . \\ \hline
\multicolumn{2}{l}{}       \\

\end{tabular}
\vspace{-8mm}
\caption{Sample translations of German sentences with rare and long words.}
\label{table:rare}
\normalsize
\end{table*}

\begin{table*}[t]
\footnotesize
\centering
\begin{tabular}{p{1.4cm}|p{14.6cm}}

\multicolumn{2}{l}{\tb{(g) Nonce words}} \\ \hline \hline
DE src  & Vieles dreht sich bei Hirst um lebensverl\"angernde oder {\clr sterbenbeschleunigende} Mittelchen . \\ \hline
EN ref  &  Much is about Hirst's life-prolonging or {\clr death-accelerating} agents .\\ \hline
bpe2char &  There are many things to do with brain around extending life-extra or {\clr accelerating} medium-sized businesses . \\ \hline
char2char & Much revolves around life extension or {\clr dying acceleration} . \\ \hline
\multicolumn{2}{l}{}       \\

\end{tabular}
\vspace{-8mm}
\caption{Sample translations of a German sentence with a nonce word.}
\label{table:nonce}
\normalsize
\end{table*}

\begin{table*}[t]
\footnotesize
\centering
\begin{tabular}{p{1.4cm}|p{14.6cm}}

\multicolumn{2}{l}{\tb{(h) Morphology}} \\ \hline \hline
DE src  & Kokainabh\"angiger Anwalt , der Drogenboss vor polizeilicher Ermittlung warnte , {\clr muss ins Gef\"angnis} . \\ \hline
EN ref  & Cocaine-addict lawyer who tipped off Mr Big about police investigation {\clr must be jailed} . \\ \hline
bpe2char & Cocaine lawyer , warning drug boss from police investigation , {\clr must in prison} . \\ \hline
char2char & Lawyers who warned the drugs boss before police investigation {\clr must be prisoners} . \\ \hline
\multicolumn{2}{l}{}       \\

\multicolumn{2}{l}{\tb{(i) Morphology}} \\ \hline \hline
DE src  & Der US-Senat genehmigte letztes Jahr ein 90 Millionen Dollar teures Pilotprojekt , das 10.000 Autos {\clr umfasst h\"atte }. \\ \hline
EN ref  &  The U.S. Senate approved a \$ 90-million pilot project last year that {\clr would have involved} about 10,000 cars . \\ \hline
bpe2char &  Last year , the US Senate approved a \$ 90 million pilot project , which {\clr would have covened} 10,000 cars . \\ \hline
char2char & The US Senate approved a \$ 90 million pilot project approved last year , which {\clr would have included} 10,000 cars . \\ \hline
\multicolumn{2}{l}{}       \\

\end{tabular}
\vspace{-8mm}
\caption{Sample translations of German sentences with morphological inflections.}
\label{table:morph}
\normalsize
\end{table*}

\begin{table*}[t]
\footnotesize
\centering
\begin{tabular}{p{1.4cm}|p{14.6cm}}

\multicolumn{2}{l}{\tb{(j) Multilingual}} \\ \hline \hline
Multi src  & \vspace{-0.85em} {\clb\foreignlanguage{russian}{Считается}} , {\clg\foreignlanguage{german}{wenn alle gut essen und sich ausreichend bewegen würden}} , {\cly\foreignlanguage{czech}{dalo by se předejít až 30 \% případů rakoviny }} .\\ \hline
EN ref  & It is estimated that {\clr if everyone ate healthily and sufficiently moved }, 30 \% of cancer cases could be prevented . \\ \hline
bpe2char &  It is considered that {\clr everyone would eat and move enough }, up to 30 \% of cases of cancer would be prevented . \\ \hline
char2char & It is considered that up to 30 \% of cancer cases could be prevented {\clr if everyone were eating well and moving sufficiently }. \\ \hline
\multicolumn{2}{l}{} \\

\multicolumn{2}{l}{\tb{(k) Multilingual}} \\ \hline \hline
Multi src  & \vspace{-0.85em} {\clb\foreignlanguage{russian}{Поиски этой технологии}} , {\cly\foreignlanguage{czech}{přivedl některé státní agentury}} {\clg\foreignlanguage{german}{zu einem kleinen Startup-Unternehmen}} {\cly\foreignlanguage{czech}{v Kalifornii s názvem True Mileage }}. \\ \hline
EN ref  & \vspace{-0.85em} {\clr The hunt for that technology} has led some state agencies to a small California startup called True Mileage . \\ \hline
bpe2char &  \vspace{-0.85em} {\clr Since this technology} has brought some State agencies to a small start-up company in California with True Mileage . \\ \hline
char2char & \vspace{-0.85em} {\clr Searching for this technology} has brought some state agencies to a small startup company in California called True Mileage . \\ \hline
\multicolumn{2}{l}{}        \\

\multicolumn{2}{l}{\tb{(l) Multilingual}} \\ \hline \hline
Multi src  & \vspace{-0.85em} {\clb\foreignlanguage{russian}{Пока американские проектировщики дорог пытаются найти деньги на ремонт рассыпающейся системы шоссейных дорог}} , {\cly\foreignlanguage{czech}{mnozí začínají vidět řešení v malé černé skříňce , která se snadno vejde do přístrojové desky vašeho auta}}.\\ \hline
EN ref  & \vspace{-0.85em} {\clr As America's road planners struggle to find the cash to mend a crumbling highway system} , many are beginning to see a solution in a little black box that fits neatly by the dashboard of your car . \\ \hline
bpe2char & \vspace{-0.85em} {\clr As long as American road designers are trying to find money for repairing the racial road system} , many are beginning to see a solution in a small black box , which will easily fit into your car's instrumentation plates . \\ \hline
char2char & \vspace{-0.85em} {\clr While American road designers are trying to find money for repairing road roads} , many are beginning to see the solution in a small black box that is easily working in the instrument plates of your car . \\ \hline
\multicolumn{2}{l}{}       \\

\end{tabular}
\vspace{-8mm}
\caption{Sample translations of mixed sentences from German (green), Czech (yellow) and Russian (blue) into English.}
\label{table:multilingual}
\normalsize
\end{table*}
*/

\end{document}